\documentclass{article} 
\usepackage{iclr2015}
\usepackage{times}
\usepackage{epsfig}
\usepackage{graphicx}
\usepackage{amsmath}
\usepackage{amssymb}
\usepackage{floatrow}

\newcommand{\cut}[1]{}

\usepackage[pagebackref=true,breaklinks=true,letterpaper=true,colorlinks,bookmarks=false]{hyperref}

\iclrfinalcopy 

\iclrconference 

\begin{document}

\title{Multiple Object Recognition with\\ Visual Attention}

\author{
\and
\textbf{Jimmy Lei Ba}\protect\footnotemark[1]\\
University of Toronto\\
{\tt\small jimmy@psi.utoronto.ca}
\and
\textbf{Volodymyr Mnih}\\
Google DeepMind\\
{\tt\small vmnih@google.com}
\and
\textbf{Koray Kavukcuoglu}\\
Google DeepMind\\
{\tt\small korayk@google.com}
}

\maketitle

\begin{abstract}
We present an attention-based model for recognizing multiple objects in images.
The proposed model is a deep recurrent neural network trained with reinforcement learning to attend to the most relevant regions of the input image.
We show that the model learns to both localize and recognize multiple objects despite being given only class labels during training.
We evaluate the model on the challenging task of transcribing house number sequences from Google Street View images and show that it is both more accurate than the state-of-the-art convolutional networks and uses fewer parameters and less computation.
\end{abstract}
\section{Introduction}
\label{sec:intro}


Convolutional neural networks have recently been very successful on a variety of
recognition and classification tasks~\citep{krizhevsky-imagenet,Goodfellow2013,Jaderberg2014synth,Vinyals2014show,Karpathy2014deep}.
One of the main drawbacks of convolutional networks (ConvNets) is their poor scalability with 
increasing input image size so efficient implementations of these models on 
multiple GPUs~\citep{krizhevsky-imagenet} or even spanning multiple machines~\citep{dean2012large}
have become necessary. 

Applications of ConvNets to multi-object and sequence recognition from images have avoided working with big images and instead focused on using ConvNets for recognizing characters or short sequence segments from image patches containing reasonably tightly cropped instances~\citep{Goodfellow2013,Jaderberg2014synth}.
Applying such a recognizer to large images containing uncropped instances requires integrating it with a separately trained sequence detector or a bottom-up proposal generator. Non-maximum suppression is often performed to obtain the final detections.
While combining separate components trained using different objective functions has been shown to be worse than end-to-end training of a single system in other domains, integrating object localization and recognition into a single globally-trainable architecture has been difficult.

In this work, we take inspiration from the way humans perform visual sequence recognition tasks such as reading by continually moving the fovea to the next relevant object or character, recognizing the individual object, and adding the recognized object to our internal representation of the sequence.
Our proposed system is a deep recurrent neural network that at each step processes a multi-resolution crop of the input image, called a \emph{glimpse}. The network uses information from the glimpse to update its internal representation of the input, and outputs the next glimpse location and possibly the next object in the sequence.
The process continues until the model decides that there are no more objects to process.
We show how the proposed system can be trained end-to-end by approximately maximizing a variational lower bound on the label sequence log-likelihood. This training procedure can be used to train the model to both localize and recognize multiple objects purely from label sequences.

We evaluate the model on the task of transcribing multi-digit house numbers from publicly available Google Street View imagery.
Our attention-based model outperforms the state-of-the-art ConvNets on tightly cropped inputs while using both fewer parameters and much less computation.
We also show that our model outperforms ConvNets by a much larger margin in the more realistic setting of larger and less tightly cropped input sequences.

\footnotetext[1]{Work done while at Google DeepMind.}
\section{Related work}

Recognizing multiple objects in images has been one of the most important goals of computer vision.
Perhaps the most common approach to image-based classification of character sequences involves combining a sliding window detector with a character classifier~\citep{Wang2012,Jaderberg2014deep}.
The detector and the classifier are typically trained separately, using different loss functions.
The seminal work on ConvNets of \cite{lecun-98} introduced a graph transformer network architecture for recognizing a sequence of digits when reading checks, and also showed how the whole system could be trained end-to-end.
That system however, still relied on a number of ad-hoc components for extracting candidate locations.

More recently, ConvNets operating on cropped sequences of characters have achieved state-of-the-art performance on house number recognition~\citep{Goodfellow2013} and natural scene text recognition~\citep{Jaderberg2014synth}.
\cite{Goodfellow2013} trained a separate ConvNets classifier for each character position in a house number with all weights except for the output layer shared among the classifiers.
\cite{Jaderberg2014synth} showed that synthetically generated images of text can be used to train ConvNets classifiers that achieve state-of-the-art text recognition performance on real-world images of cropped text.

Our work builds on the long line of the previous attempts on attention-based visual processing~\citep{Itti:PAMI:1998,Larochelle:NIPS:2010,Alexe:NIPS:2012}, and in particular extends the recurrent attention model (RAM) proposed in~\cite{Mnih2014}.
While RAM was shown to learn successful gaze strategies on cluttered digit classification tasks and on a toy visual control problem it was not shown to scale to real-world image tasks or multiple objects.
Our approach of learning by maximizing variational lower bound is equivalent to the reinforcement learning procedure used in RAM and is related to the work of \cite{Maes2009} who showed how reinforcement learning can be used to tackle general structured prediction problems.





\begin{figure*}[t]%
    \label{fig:dram}
    \centering
    \vspace{-0.5in}
    \includegraphics[height=2.4in]{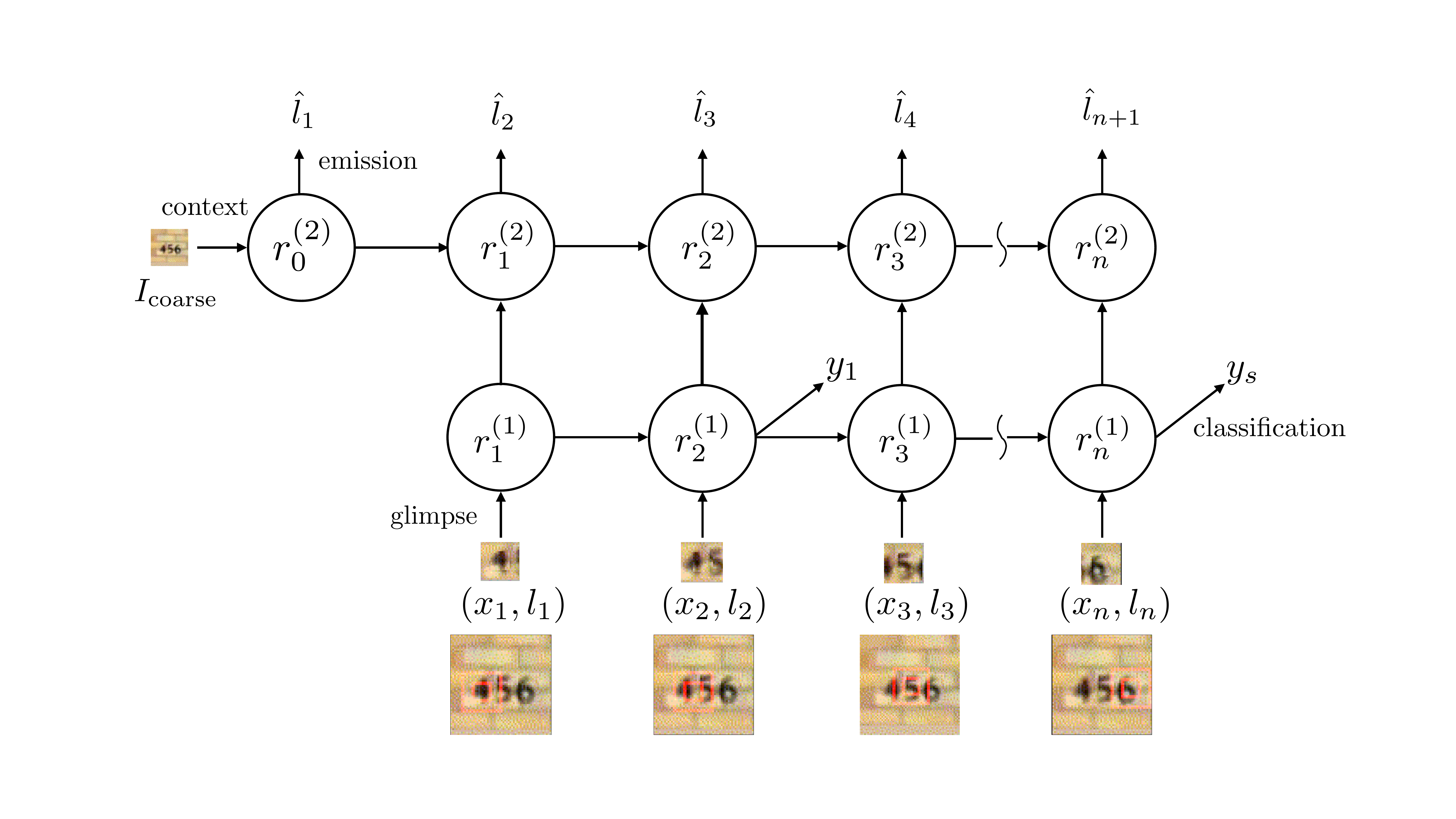}
    ~\\[-.25in]
    \caption{The deep recurrent attention model.} 
\end{figure*}

\section{Deep recurrent visual attention model}
\label{sec:method}

For simplicity, we first describe how our model can be applied to classifying a single object and later show how it can be extended to multiple objects.
Processing an image $x$ with an attention-based model is a sequential process with $N$ steps, where each
step consists of a saccade followed by a glimpse.
At each step $n$, the model receives a location $l_n$ along with a glimpse observation $x_n$ taken at location $l_n$.
The model uses the observation to update its internal state and outputs the location $l_{n+1}$ to process at the next time-step.
Usually the number of pixels in the glimpse $x_n$ is much smaller than the number of pixels in the original
image $x$, making the computational cost of processing a single glimpse independent of the size of the image.



A graphical representation of our model is shown in Figure~\ref{fig:dram}.
The model can be broken down into a number of sub-components, each mapping some input into a vector output.
We will use the term ``network'' to describe these non-linear sub-components since they are typically
multi-layered neural networks.

\textbf{Glimpse network}: The glimpse network is a non-linear function that receives
the current input image patch, or glimpse, $x_n$ and its location tuple $l_n$ , where $l_n = (x_n, y_n)$, as input
and outputs a vector $g_n$.
The job of the glimpse network is to extract a set of useful features from location $l_n$ of the raw visual input.
We will use
$G_{image}(x_n|W_{image})$ to denote the output vector from function
$G_{image}(\cdot)$ that takes an image patch $x_n$ and is parameterized by weights $W_{image}$. 
$G_{image}(\cdot)$ typically consists of three convolutional hidden layers
without any pooling layers followed by a fully connected layer.
Separately, the location tuple is mapped by $G_{loc}(l_n|W_{loc})$
using a fully connected hidden layer where, both $G_{image}(x_n|W_{image})$ and
$G_{loc}(l_n|W_{loc})$ have the same dimension.
We combine the high bandwidth image information with the low bandwidth location tuple
 by multiplying the two
vectors element-wise to get the final glimpse feature vector $g_n$,
\begin{align}
    g_n = G_{image}(x_n|W_{image})G_{loc}(l_n|W_{loc}).
\end{align} 
This type of multiplicative interaction between ``what'' and ``where'' was initially proposed by~\cite{Larochelle:NIPS:2010}.

\textbf{Recurrent network}: The recurrent network aggregates information extracted 
from the
individual glimpses and combines the information in a coherent manner that preserves spatial information.
The glimpse feature vector $g_n$ from the glimpse network is supplied as input
to the recurrent network at each time step. The recurrent network consists of two
recurrent layers with non-linear function $R_{recur}$. We defined the two outputs of the recurrent layers as $r^{(1)}$ and
$r^{(2)}$.
\begin{align}
    r^{(1)}_n = R_{recur}(g_n, r^{(1)}_{n-1}|W_{r1}) \text{ and }
    r^{(2)}_n = R_{recur}(r^{(1)}_n, r^{(2)}_{n-1}|W_{r2})
\end{align}
We use Long-Short-Term Memory units~\citep{hochreiter1997lstm} for the non-linearity $R_{recur}$ because of their ability to learn long-range dependencies and stable learning dynamics.

\textbf{Emission network}: The emission network takes the current state of recurrent network
as input and makes a prediction on where to extract the
next image patch for the glimpse network. It acts as a controller that directs 
attention based on the current internal states from 
the recurrent network. It consists of a fully connected
hidden layer that maps the feature vector $r^{(2)}_n$ from the top recurrent layer to
a coordinate tuple $\hat{l}_{n+1}$. 
\begin{align}
    \hat{l}_{n+1} = E(r_n^{(2)}|W_{e})
\end{align}

\textbf{Context network}: The context network provides the initial state for the
recurrent network and its output is used by the emission network to predict the
location of the first glimpse. The context network
$C(\cdot)$ takes a down-sampled low-resolution version of the whole input image
$I_{coarse}$ and outputs a fixed length vector $c_{I}$. The contextual
information provides sensible hints on where the potentially interesting
regions are in a given image. The context network employs three
convolutional layers that map a coarse image $I_{coarse}$ to a feature vector
used as the initial state of the top recurrent layer $r^{2}$ in the recurrent
network. However, the bottom layer $r^{1}$ is initialized with a vector of zeros for reasons we will explain later.

\textbf{Classification network}: 
The classification network outputs a prediction for the class label $y$ based on the final feature vector $r_N^{(1)}$ of the lower recurrent layer.
The classification network has one fully connected hidden layer and a softmax output layer for the class $y$.
\begin{align}
    P(y|I) = O(r^{1}_n|W_o)
\end{align} 

Ideally, the deep recurrent attention model should learn to look at locations that are relevant for classifying objects of interest. 
The existence of the contextual information, however, provides a
``short cut'' solution such that it is much easier for the model to learn from
contextual information than by combining information from different glimpses.
We prevent such
undesirable behavior by connecting the context network and classification
network to different recurrent layers in our deep model. As a result, the contextual
information cannot be used directly by the classification network and only affects the sequence of glimpse locations produced by the model.


\subsection{Learning where and what}

Given the class labels $y$ of image $I$, we can formulate learning as a supervised
classification problem with the cross entropy objective function. The attention model
predicts the class label conditioned on intermediate latent location variables
$l$ from each glimpse and extracts the corresponding patches. We can thus 
maximize likelihood of the class label by marginalizing over the glimpse locations
$\log p(y|I,W) = \log \sum_l p(l|I,W)p(y|l,I,W)$.

The marginalized objective function can be learned through optimizing its 
variational free energy lower bound $\mathcal{F}$:
\begin{align}
    \log \sum_l p(l|I,W)p(y|l,I,W) &\ge \sum_l p(l|I,W) \log p(y,l|I,W) + H[l] \label{eq:obj}\\
    &= \sum_l p(l|I,W) \log p(y|l,I,W)
    \label{eq:free_energy}
\end{align}

The learning rules can be derived by taking derivatives
of the above free energy with respect to the model parameter $W$:
\begin{align}
    {\partial \mathcal{F} \over \partial W} &= \sum_l p(l|I,W) {\partial \log p(y|l,I,W) \over \partial W} +  \sum_l \log p(y|l,I,W){\partial p(l|I,W) \over \partial W} \\
    &= \sum_l p(l|I,W) \bigg[ {\partial \log p(y|l,I,W) \over \partial W} + \log p(y|l,I,W){\partial \log p(l|I,W) \over \partial W} \bigg]
    \label{eq:derivatives}
\end{align}

For each glimpse in the glimpse sequence, it is difficult to evaluate
exponentially many glimpse locations during training.  The summation in
equation \ref{eq:derivatives} can then be approximated using Monte Carlo
samples.  
\begin{align}
    \tilde{l}^m \sim p(l_n|I,W) &= \mathcal{N}(l_n; \hat{l}_n, \Sigma)\label{eq:gaussian} \\
    {\partial \mathcal{F} \over \partial W} \approx {1\over M}\sum_{m=1}^M \bigg[ {\partial \log p(y|\tilde{l}^m,I,W) \over \partial W}& + \log p(y|\tilde{l}^m,I,W){\partial \log p(\tilde{l}^m|I,W) \over \partial W} \bigg]
    \label{eq:mc_approx}
\end{align}

The equation \ref{eq:mc_approx} gives a practical algorithm to train the deep
attention model. Namely, we can sample the glimpse location prediction from the
model after each glimpse. The samples are then used in the standard
backpropagation to obtain an estimator for the gradient of the model
parameters. Notice that log likelihood $\log p(y|\tilde{l}^m, I,W)$ has an
unbounded range that can introduce substantial high variance in the gradient 
estimator.  Especially when the sampled location is off from the object in the
image, the log likelihood will induce an undesired large gradient update that is 
backpropagated through the rest of the model. 

We can reduce the variance in the estimator \ref{eq:mc_approx} by replacing the
$\log p(y|\tilde{l}^m, I,W)$ with a 0/1 discrete indicator function $R$ and
using a baseline technique used in \cite{Mnih2014}.
\begin{align}
    R = 
         \begin{cases} 
          1 & y = \arg\max_y \log p(y|\tilde{l}^m, I, W) \\
          0 & \text{otherwise}
         \end{cases}
    \label{eq:reward}
\end{align}
\begin{align}
    b_n = E_{baseline}(r_n^{(2)}|W_{baseline})
    \label{eq:baseline}
\end{align} 

As shown, the recurrent network state vector $r_n^{(2)}$ is used to estimate a
state-based baseline $b$ for each glimpse that significantly improve the
learning efficiency. The baseline effectively centers the random variable $R$
and can be learned by regressing towards the expected value of $R$. Given both
the indicator function and the baseline, we have the following gradient update:
\begin{align}
    {\partial \mathcal{F} \over \partial W} \approx {1\over M}\sum_{m=1}^M \bigg[ {\partial \log p(y|\tilde{l}^m,I,W) \over \partial W}& + \lambda(R - b){\partial \log p(\tilde{l}^m|I,W) \over \partial W} \bigg]
    \label{eq:reinforce}
\end{align} 
where, hyper-parameter $\lambda$ balances the scale of the two gradient
components. In fact, by using the 0/1 indicator function, the learning rule from
equation \ref{eq:reinforce} is equivalent to the REINFORCE~\citep{Williams1992} learning rule
employed in \cite{Mnih2014} for training their attention model.  When viewed as a reinforcement learning update, the second term in equation~\ref{eq:reinforce} is an
unbiased estimate of the gradient with respect to $W$ of the expected reward $R$ under the model glimpse policy. Here we show that such
learning rule can also be motivated by simply approximately optimizing the free energy. 

During inference, the feedforward location prediction can be used as a
deterministic prediction on the location coordinates to extract the next input
image patch for the model. The model behaves as a normal feedforward network.
Alternatively, our marginalized objective function equation \ref{eq:obj}
suggests a procedure to estimate the expected class prediction by using samples of
location sequences $\{{\tilde{l}}_1^m, \cdots, {\tilde{l}}_N^m\}$ and averaging
their predictions,
\begin{align}
    \mathbb{E}_{l}[p(y|I)] \approx {1\over M}\sum_{m=1}^M p(y|I, \tilde{l}^m).
\end{align}
This allows the attention model to be evaluated multiple times on
each image with the classification predictions being averaged. In
practice, we found that averaging the log probabilities gave the best performance.

In this paper, we encode the real valued glimpse location
tuple $l_n$ using a Cartesian coordinate that is centered at the middle of the
input image. The ratio converting unit width in the coordinate system to the number of pixels
is a hyper-parameter. This ratio presents an exploration versus exploitation trade off.
The proposed model performance is very sensitive to this setting. We found that 
setting its value to be around 15\% of the input image width tends to work well.

\subsection{Multi-object/Sequential classification as a visual attention task}

Our proposed attention model can be easily extended to solve classification
tasks involving multiple objects. To train the deep recurrent attention model for
the sequential recognition task, the multiple object labels for a given image need
to be cast into an ordered sequence $\{y_1, y_2, \cdots, y_s\}$. The deep recurrent
attention model then learns to predict one object at a time as it explores the
image in a sequential manner. We can utilize a simple fixed number of glimpses
for each target in the sequence. In addition, a new class label for the
``end-of-sequence'' symbol is included to deal with variable numbers of objects in an
image. We can stop the recurrent attention model once a terminal symbol is
predicted. Concretely, the objective function for the sequential prediction 
is 
\begin{align}
\log p(y_1, y_2, \cdots, y_S|I,W) = \sum_{s=1}^S \log \sum_l p(l_s|I,W) p(y_s |l_s, I, W)
\label{eq:seq_obj}
\end{align}
The learning rule is derived as in equation
\ref{eq:reinforce} from the free energy and the gradient is accumulated across all targets. We assign a fixed
number of glimpses, $N$, for each target. Assuming $S$ targets in an
image, the model would be trained with $N\times (S+1)$ glimpses.  The
benefit of using a recurrent model for multiple object recognition is that it
is a compact and simple form yet flexible enough to deal with images containing
variable numbers of objects.

Learning a model from images of many objects is a challenging setup.
We can reduce the difficulty by modifying our indicator function $R$ to be proportional 
to the number of targets the model predicted correctly.  
\begin{align}
    R_s = \sum_{j\le s} R_j 
    \label{eq:seq_reward}
\end{align}
In addition, we restrict the gradient of the objective function so that it only
contains glimpses up to the first mislabeled target and ignores the targets after
the first mistake. This curriculum-like adaption to the learning is crucial to
obtain a high performance attention model for sequential prediction.

\section{Experiments}

To show the effectiveness of the deep recurrent attention model (DRAM), we first investigate
a number of multi-object classification tasks involving a variant of MNIST. We then apply
the proposed attention model to a real-world object recognition task using the
multi-digit SVHN dataset \cite{netzer2011reading} and compare with the
state-of-the-art deep ConvNets.  A description of the models and training protocols we used can be found in the Appendix.

As suggested in \cite{Mnih2014}, classification performance can be improved by having
a glimpse network with two different scales. Namely, given a glimpse location
$l_n$, we extract two patches $(x^{1}_n, x^{2}_n)$ where $x^{1}_n$ is the
original patch and $x^{2}_n$ is a down-sampled coarser image patch.  We use the
concatenation of $x_n^1$ and $x_n^2$ as the glimpse observation.
``foveal'' feature. 

The hyper-parameters in our experiments are the learning rate $\eta$ and the
location variance $\Sigma$ in equation \ref{eq:gaussian}. They are determined
by grid search and cross-validation.

\subsection{Learning to find digits}

We first evaluate the effectiveness of the controller in the deep recurrent
attention model using the MNIST handwritten digit dataset.  

We generated a dataset of pairs of randomly picked handwritten digits in a 100x100
image with distraction noise in the background.  The task is to identify the 55
different combinations of the two digits as a classification problem. The
attention models are allowed 4 glimpses before making a classification
prediction.
The goal of this
experiment is to evaluate the ability of the controller and recurrent network
to combine information from multiple glimpses with minimum effort from the
glimpse network. The results are shown in table (\ref{tb:2MNIST}).  The DRAM
model with a context network significantly outperforms the other models. 

\begin{figure}
\begin{floatrow}%
   \ttabbox{
   \begin{tabular}{|l|c|} \hline
           {Model} & {Test Err.} \\ \hline \hline
           RAM \cite{Mnih2014} & 9\% \\\hline
           DRAM w/o context & {7\%} \\\hline
           DRAM & \bf{{5\%}} \\\hline
   \end{tabular} 
   \label{tb:2MNIST}
   }
   {\caption{Error rates on the MNIST pairs classification task. }} 
   \ttabbox{
    \begin{tabular}{|l|c|} \hline
           {Model} & {Test Err.} \\ \hline \hline
           ConvNet 64-64-64-512 & 3.2\% \\\hline
           DRAM & \bf{{2.5\%}} \\\hline
    \end{tabular} 
   \label{tb:2MNIST_add} 
   }
   {\caption{Error rates on the MNIST two digit addition task.}}
   \end{floatrow}
\end{figure}

\subsection{Learning to do addition}

For a more challenging task, we designed another dataset with two MNIST digits on an empty 100x100 background where the task
is to predict the sum of the two digits in the image as a classification problem with 19 targets. The model
has to find where each digit is and add them up.  When the two digits are
sampled uniformly from all classes, the label distribution is heavily
imbalanced for the summation where most of the probability mass concentrated
around 10. Also, there are many digit combinations that can be mapped to the
same target, for example, [5,5] and [3,7].

The class label provides a weaker association between the visual feature and
supervision signal in this task than in the digit combination task.  We used
the same model as in the combination task. The deep recurrent attention model
is able to discover a glimpse policy to solve this task achieving a 2.5\%
error rate. In comparison, the ConvNets take longer to learn and perform
worse when given weak supervision.
\begin{figure*}[t]%
    \centering
    \vspace{0.in}
    \includegraphics[height=1.015in]{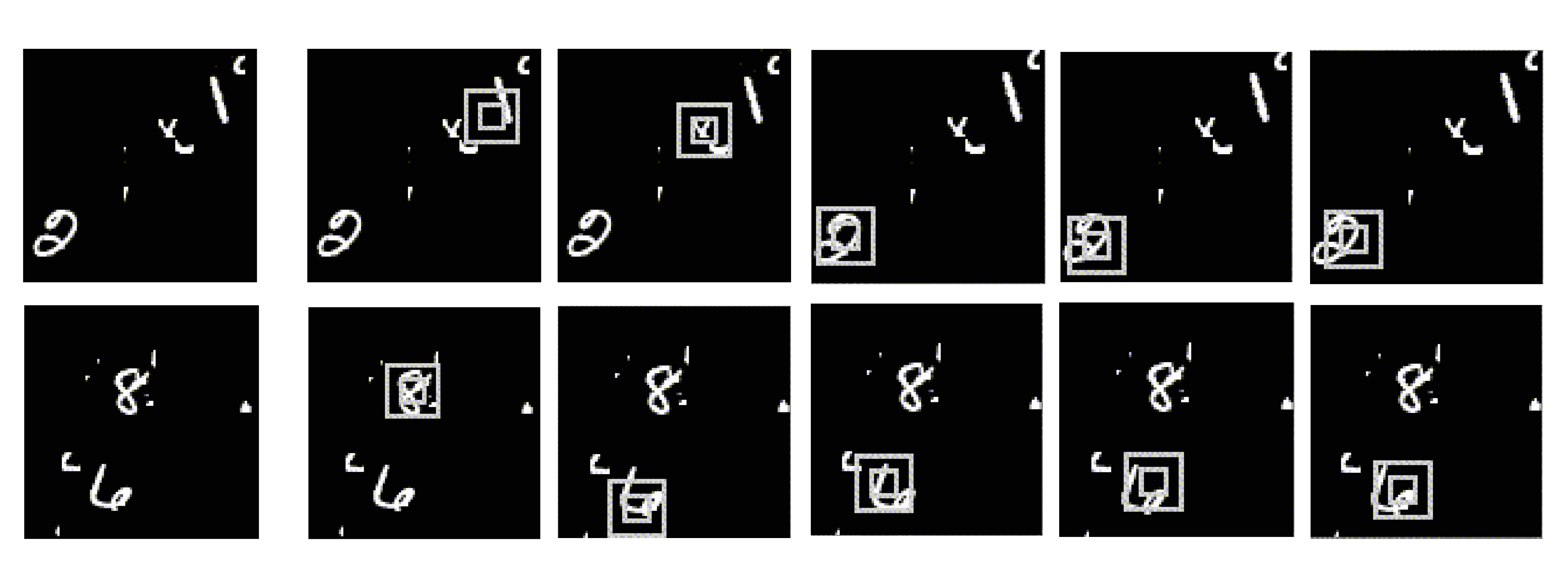}
    \includegraphics[height=1.015in]{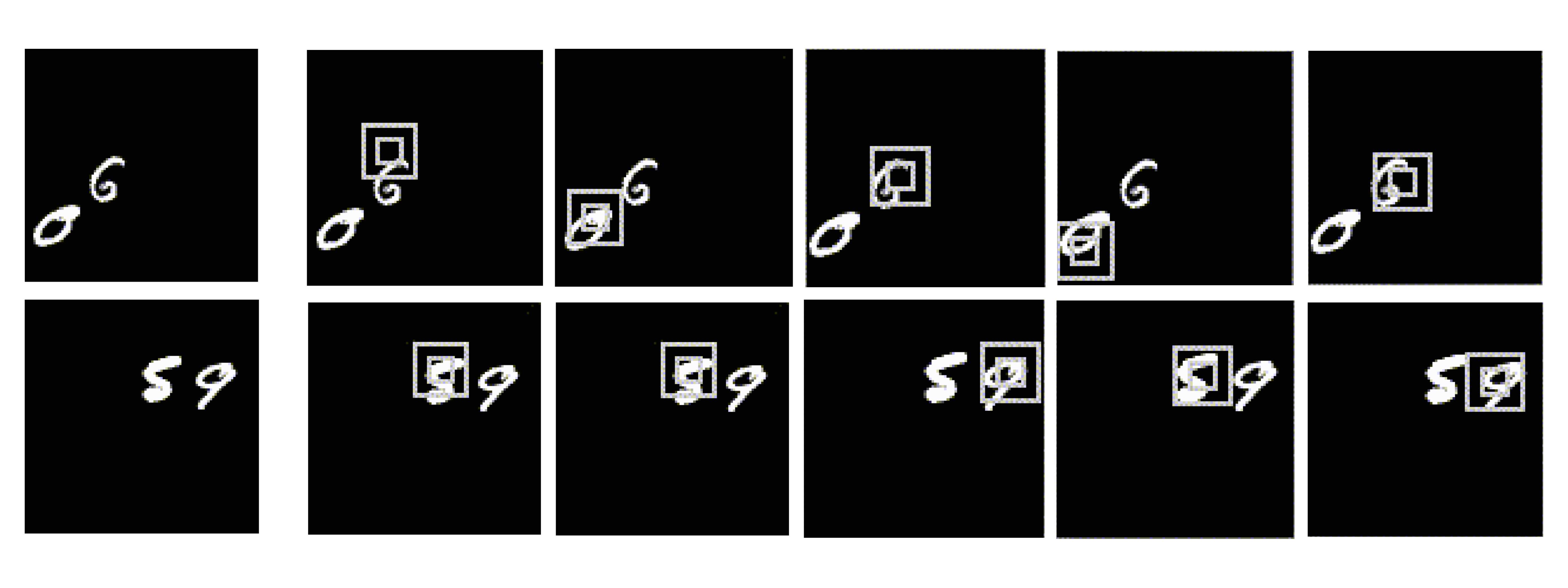}
    \caption{\textbf{Left)} Two examples of the learned policy on the digit pair classification task.  The first column shows the input image while the next 5 columns show the selected glimpse locations. \textbf{Right)} Two examples of the learned policy on the digit addition task.  The first column shows the input image while the next 5 columns show the selected glimpse locations. } 
    \label{fig:dram_2mnist_combine}
\end{figure*}

Some inference samples are shown in figure~\ref{fig:dram_2mnist_combine} It is
surprising that the learned glimpses policy for predicting the next glimpse is
very different in the addition task comparing to the predicting combination
task.  The model that learned to do addition toggles its glimpses between the
two digits.  

\subsection{Learning to read house numbers}
\begin{figure}
\begin{floatrow}%
   \ttabbox{
   \begin{tabular}{|l|c|} \hline
           {Model} & {Test Err.} \\ \hline \hline
           11 layer CNN \cite{Goodfellow2013} & 3.96\% \\\hline
           10 layer CNN & 4.11\% \\\hline
           Single DRAM & {{5.1\%}} \\\hline
           Single DRAM MC avg. & {{4.4\%}} \\\hline
           forward-backward DRAM MC avg. & {\bf{3.9\%}} \\\hline
    \end{tabular}    
    \label{tb:svhn} 
   }
   {\caption{Whole sequence recognition error rates on multi-digit SVHN.}} 
   \ttabbox{
    \begin{tabular}{|l|c|} \hline
           {Model} & {Test Err.} \\ \hline \hline
           10 layer CNN resize & 50\% \\\hline
           10 layer CNN re-trained & 5.60\% \\\hline
           Single DRAM focus & {{5.7\%}} \\\hline
           forward-backward DRAM focus & {{5.0\%}} \\\hline
           Single DRAM fine-tuned & {{5.1\%}} \\\hline
           forward-backward DRAM fine-tuning & {\bf{4.46\%}} \\\hline
    \end{tabular} 
    \label{tb:svhn_enlarge} 
   }
   {\caption{Whole sequence recognitionn error rate on enlarged multi-digit SVHN.}}
   \end{floatrow}
\end{figure}

The publicly available multi-digit street view house number (SVHN) dataset
\cite{netzer2011reading} consists of images of digits taken from pictures of
house fronts.
Following \cite{Goodfellow2013}, we
formed a validation set of 5000 images by randomly sampling images from the training set and the extra set, and these were used for selecting the
learning rate and sampling variance for the stochastic glimpse policy.  The models are
trained using the remaining 200,000 training images. We follow the
preprocessing technique from \cite {Goodfellow2013} to generate tightly cropped 
64 x 64 images with multi-digits at the center and similar data augmentation is
used to create 54x54 jittered images during training. We also convert the RGB
images to grayscale as we observe the color information does not affect
the final classification performance.

We trained a model to classify all the digits in an image sequentially with the
objective function defined in equation \ref{eq:seq_obj}. The label sequence ordering is chosen to 
go from left to right as the natural ordering of the house number. The attention model
is given 3 glimpses for each digit before making a prediction. The
recurrent model keeps running until it predicts a terminal label or until the
longest digit length in the dataset is reached.  In the SVHN dataset,
up to 5 digits can appear in an image. This means the recurrent model will run
up to 18 glimpses per image, that is 5 x 3 plus 3 glimpses for a terminal
label. Learning the attention model took around 3 days on a GPU. 

The model performance is shown in table (\ref{tb:svhn}). We found that there is
still a performance gap between the state-of-the-art deep ConvNet
and a single DRAM that ``reads'' from left to right, even with the Monte
Carlo averaging.  The DRAM often over predicts additional digits in the place
of the terminal class.  In addition, the distribution of the leading digit in
real-life follows Benford's law. 

We therefore train a second recurrent attention model to ``read'' the house
numbers from right to left as a backward model. The forward and backward model
can share the same weights for their glimpse networks but they have different
weights for their recurrent and their emission networks. The predictions of
both forward and backward models can be combined to estimate the final sequence
prediction. Following the observation that attention models often overestimate
the sequence length, we can flip first $k$ number of sequence prediction from
the backwards model, where $k$ is the shorter length of the sequence length
prediction between the forward and backward model. This simple heuristic works
very well in practice and we obtain state-of-the-art performance on the Street
View house number dataset with the forward-backward recurrent attention model.
Videos showing sample runs of the forward and backward models on SVHN test data
can be found at \url{http://www.psi.toronto.edu/~jimmy/dram/forward.avi} and
\url{http://www.psi.toronto.edu/~jimmy/dram/backward.avi} respectively.  These
visualizations show that the attention model learns to follow the slope of
multi-digit house numbers when they go up or down.

For comparison, we also implemented a deep ConvNet with a similar architecture to the one used in \cite{Goodfellow2013}. The network had 8 convolutional layers with 128
filters in each followed by 2 fully connected layers of 3096 ReLU units.
Dropout is applied to all 10 layers with 50\% dropout rate to prevent
over-fitting.   
\begin{table*}%
    \center
    \begin{tabular}{|l|c|c|c|c|} \hline
            (Giga) floating-point op. & 10 layer CNN & DRAM & DRAM MC avg. & F-B DRAM MC avg. \\ \hline \hline
           54x54  & 2.1 & $\le$0.2 & 0.35 & 0.7 \\\hline
           110x110  & 8.5 & $\le$0.2 & 1.1 & 2.2 \\\hline
    \end{tabular} 
   \begin{tabular}{|l|c|c|c|c|} \hline
            param. (millions)  & 10 layer CNN & DRAM & DRAM MC avg. & F-B DRAM MC avg. \\ \hline \hline
           54x54 & 51 &  14 & 14 & 28 \\\hline
           110x110 & 169 &  14 & 14 & 28 \\\hline
    \end{tabular} 
   \caption{Computation cost of DRAM V.S. deep ConvNets} 
   \label{tb:svhn_flops} 
\end{table*}

Moreover, we generate a less cropped 110x110 multi-digit SVHN dataset by
enlarging the bounding box of each image such that the relative size of the
digits stays the same as in the 54x54 images. Our deep attention
model trained on 54x54 can be directly applied to the new 110x110 dataset with
no modification. The performance can be further improved
by ``focusing'' the model on where the digits are. We run the model once
and crop a 54x54 bounding box around the glimpse location sequence and feed the
54x54 bounding box to the attention model again to generate the final
prediction. This allows DRAM to ``focus'' and obtain a similar prediction accuracy on the
enlarged images as on the cropped image without ever being trained on large images. 
We also compared the deep ConvNet trained on the 110x110
images with the fine tuned attention model.  The deep attention model significantly
outperforms the deep ConvNet with very little training time. The DRAM
model only takes a few hours to fine-tune on the enlarged SVHN data, compared to one week
for the deep 10 layer ConvNet. 

\section{Discussion}

In our experiments, the proposed deep recurrent attention model (DRAM) outperforms 
the state-of-the-art deep ConvNets on the standard SVHN sequence recognition task.
Moreover, as we increase the image area around the house numbers or lower the 
signal-to-noise ratio, the advantage of the attention model becomes more significant.

In table~\ref{tb:svhn_flops}, we compare the computational cost of our proposed deep 
recurrent attention model with that of deep ConvNets in terms of the number 
of float-pointing operations for the multi-digit SVHN models along with the number of 
parameters in each model. The recurrent attention models that only process a selected 
subset of the input scales better than a ConvNet that looks over an entire image. The estimated
cost for the DRAM is calculated using the maximum sequence length in the
dataset, however the expected computational cost is much lower in practice since most of
the house numbers are around $2-3$ digits long.  In addition, since the attention based model does not 
process the whole image, it can naturally work on images of different size with the same 
computational cost independent of the input dimensionality.

We also found that the attention-based model is less prone to
over-fitting than ConvNets, likely because of the stochasticity 
in the glimpse policy during training.  Though it is still beneficial to 
regularize the attention model with some dropout noise between the hidden 
layers during training, we found that it gives a very marginal performance 
boost of 0.1\% on the multi-digit SVHN task.  On the other hand, the deep 
10 layer ConvNet is only able to achieve 5.5\% error rate when dropout 
is only applied to the last two fully connected hidden layer.

Finally, we note that DRAM can easily deal with variable length label 
sequences. Moreover, a model trained on a dataset with a fixed sequence length 
can easily be transferred and fine tuned with a similar dataset but longer 
target sequences. This is especially useful when there is lack of data for 
the task with longer sequences.

\section{Conclusion}

We described a novel computer vision model that uses an attention mechanism
to decide where to focus its computation and showed how it can be trained end-to-end
to sequentially classify multiple objects in an image.
The model outperformed the state-of-the-art ConvNets on a multi-digit house number recognition task while using both fewer parameters and less computation than the best ConvNets, thereby showing that
attention mechanisms can improve both the accuracy and efficiency of ConvNets on a real-world task.
Since our proposed deep recurrent attention model is flexible, powerful, and efficient, we believe that it may be a promising approach for tackling other challenging computer vision tasks.

\section{Acknowledgements}

We would like to thank Geoffrey Hinton, Nando de Freitas and Chris Summerfield for many helpful comments and discussions.
We would also like to thank the developers of DistBelief~\citep{dean2012distbelief}.

{\small
\bibliographystyle{iclr2015}
\bibliography{dram}
}

\section{Appendix}

\subsection{General Training Details} 
We used the ReLU activation function in the hidden layers, $g(x) = \max(0, x)$,
for the rest of the results reported here or otherwise noted. We found that
ReLU units significantly speed up training.  We optimized the model parameters
using stochastic gradient descent with the Nesterov momentum technique. A
mini-batch size of 128 was used to estimate the gradient direction. The momentum
coefficient was set to $0.9$ throughout the training. The learning rate $\eta$
scheduling was applied in training to improve the convergence of the learning
process. $\eta$ starts at $0.01$ in the first epoch and was exponentially
reduced by a factor of $0.97$ after each epoch.

\subsection{Details of Learning to Find Digits}
The unit width for the Cartesian coordinates was set to 20 and
glimpse location sampling standard deviation was set to 0.03. There are 512 LSTM
units and 256 hidden units in each fully connected layer of the model. We
intentionally used a simple fully connected single hidden layer network of 256
hidden units as $G_{image}(\cdot)$ in the glimpse network.

\subsection{Details of Learning to Read House Numbers}
Unlike in the MNIST experiment, the number of digits in each image varies and digits
have more variations due to natural backgrounds, lighting variation, and highly
variable resolution. We use a much larger deep recurrent attention model for
this task. It was crucial to have a powerful glimpse network to obtain
good performance. As described in section 3, the glimpse network consists of
three convolutional layers with 5x5 filter kernels in the first layer and 3x3
in the later two.  The number of filters in those layers was \{64, 64,
128\}.There are 512 LSTM units in each layer of the recurrent network. Also,
the fully connected hidden layers all have 1024 ReLU hidden units in each
module listed in section 3. The Cartesian coordinate unit width was set to 12 pixels
and glimpse location is sampled from a fixed variance of 0.03.

\begin{table}%
    \center
    \begin{tabular}{|l|c|} \hline
           {Model} & {Test Err.} \\ \hline \hline
           small DRAM & {{5.1\%}} \\\hline
           small DRAM + dropout & {{4.6\%}} \\\hline
    \end{tabular} 
   \caption{Effectiveness of Regularization} 
   \label{tb:svhn_reg} 
\end{table}

\end{document}